\documentclass[12pt]{mytmp}

\usepackage{fullpage}

\usepackage[utf8]{inputenc} 
\usepackage[T1]{fontenc}    
\usepackage{hyperref}       
\usepackage{url}            
\usepackage{booktabs}       
\usepackage{amsfonts}       
\usepackage{nicefrac}       
\usepackage{microtype}      
\usepackage{color}
\usepackage{graphicx}
\usepackage{multirow}

\newcommand{\tabincell}[2]{\begin{tabular}{@{}#1@{}}#2\end{tabular}}

\def\AIHC{{\sc{TStarBot2}}}
\def\AIRL{{\sc{TStarBot1}}}

\journal{Artificial Intelligence}

\begin{document}

\begin{frontmatter}

\title{{\sc TStarBots}: Defeating the Cheating Level Builtin AI in StarCraft II in the Full Game}

\author[ailab]{Peng Sun\corref{cor1}}
\author[ailab]{Xinghai Sun\corref{cor1}}
\author[ailab]{Lei Han\corref{cor1}}
\author[ailab]{Jiechao Xiong\corref{cor1}}
\author[ailab]{Qing Wang}
\author[ailab]{Bo Li}
\author[ailab]{Yang Zheng}
\author[ailab,ur]{Ji Liu}
\author[ailab]{Yongsheng Liu}
\author[ailab,nu]{Han Liu}
\author[ailab]{Tong Zhang}

\cortext[cor1]{Equal contribution}

\address[ailab]{Tencent AI Lab, China}
\address[ur]{University of Rochester, USA}
\address[nu]{Northwestern University, USA}

\begin{abstract}
Starcraft II (SC2) is widely considered as the most challenging Real
Time Strategy (RTS) game. The underlying challenges include a large observation
space, a huge (continuous and infinite) action space, 
partial observations, simultaneous move for all players, and
long horizon delayed rewards for local decisions.
To push the frontier of AI research, Deepmind and Blizzard jointly
developed the StarCraft II Learning Environment (SC2LE) as a testbench
of complex decision making systems.
SC2LE provides a few mini games such as \textit{MoveToBeacon},
\textit{CollectMineralShards}, and \textit{DefeatRoaches}, where some
AI agents have achieved the performance level of human professional players. 
However, for \emph{full} games, the current AI agents are still far
from achieving human professional level performance.
To bridge this gap, we present two full game AI agents in this paper ---
the AI agent~\AIRL~is based on deep reinforcement learning over a flat
action structure, 
and the AI agent~\AIHC~is based on hard-coded rules over a hierarchical action structure. 
Both \AIRL~and \AIHC~are able to defeat the built-in AI agents from
level 1 to level 10 in a full game (1v1 \textit{Zerg}-vs-\textit{Zerg}
game on the AbyssalReef map), noting that level 8, level 9, and level
10 are cheating agents with unfair advantages such as full vision on the whole map and resource harvest boosting
\footnote{According to some informal discussions from the StarCraft II forum, level 10 built-in AI is estimated to be Platinum to Diamond~\cite{scii-forum}, which are equivalent to top 50\% - 30\% human players in the ranking system of Battle.net Leagues~\cite{liquid}. 
}.
To the best of our knowledge, this is the first public work to investigate AI agents that can defeat the built-in AI in the StarCraft II full game. 
\end{abstract}

\end{frontmatter}

\section{Introduction}
\label{sec:intro}
Recently, the marriage of Deep Learning~\cite{goodfellow2016deep} and
Reinforcement Learning (RL)~\cite{sutton1998introduction} leads to
significant breakthroughs in machine based decision making systems,
especially for computer games. 
Systems based on deep reinforcement learning (DRL), trained either from scratch or from a
pre-trained model, can
take inputs of raw observation features and achieve impressive performance in a wide range of applications, 
including playing the board game GO~\cite{silver2016mastering, silver2017mastering}, 
playing video games (e.g., Atari~\cite{mnih2015human}, the first person shooting game Doom/ViZDoom~\cite{kempka2016vizdoom, wu2017} or Quake/DeepmindLab~\cite{beattie2016deepmind}, Dota 2~\cite{openai-five}), Robot Visuomotor Control~\cite{levine2016end}, Robot Navigation~\cite{zhu2016target, sadeghi2016cad}, etc. 
The learned policy/controller can work surprisingly well,
and in many cases even achieves super-human performance \cite{mnih2015human, silver2017mastering}.

\begin{figure}[t]
\center
\includegraphics[width=0.8\linewidth]{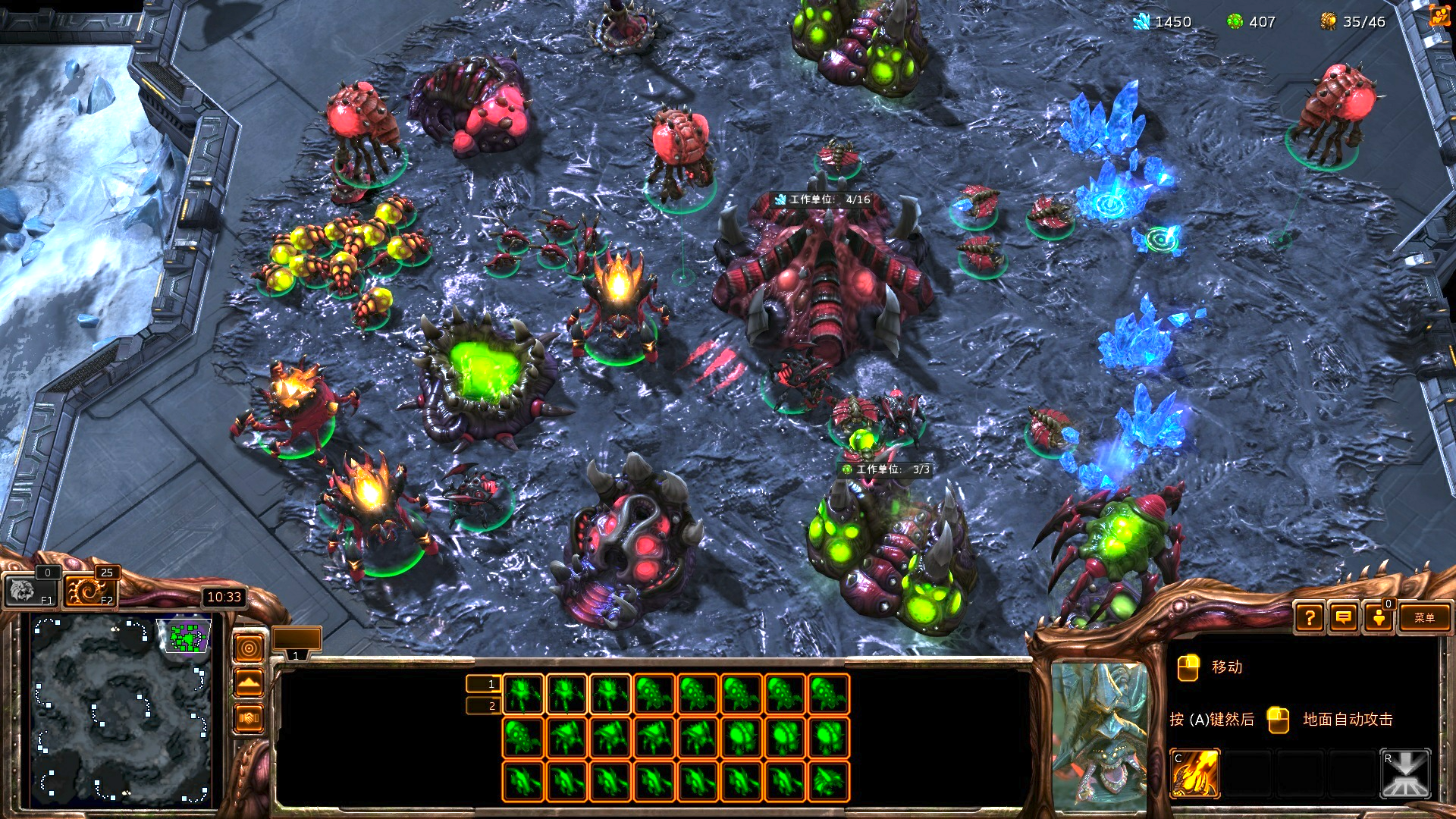}
\caption{A screenshot of the SC2 game when a human player is manipulating units.}
\label{fig:sc2game}
\end{figure}

However, Starcraft II (SC2) \cite{sc2le}, which is widely considered as the most challenging RTS game, still remains unsolved. In SC2, 
a human player has to manipulate tens to hundreds of units\footnote{In the 1 vs 1 game, the maximal number of moving units controlled by a player could exceed a hundred.
} for multiple purposes,
e.g., collecting two types of resources, expanding for extra
resources, upgrading technologies, building other units, sending
squads for attacking or defending, performing micro managements over
each unit for a battle, etc. This is one important factor why SC2 is
more challenging than Dota 2, in which the total number of units
needed to be controlled is up to five (manipulated by five players respectively).
Figure~\ref{fig:sc2game} shows a screenshot of what human players work with. 
The units of the opponent are hidden to the player, unless they are in
the viewing range of the units controlled by the player.
The player needs to send scouting units to spy
on the opponent's strategy. 
All decisions must be made in real time. 
In terms of designing AI agents,
SC2 involves a large observation space, a huge action space, partial observations, simultaneous move for all players, and
long horizon delayed rewards for local decisions.
All of these factors make SC2 extremely challenging. 
To push the frontier of AI research, Deepmind and Blizzard jointly developed the StarCraft II Learning Environment (SC2LE) \cite{sc2le}.
Some recent results by Deepmind \cite{sc2le, zambaldi2018relational}
showed that their AI agent can achieve the performance level of professional players in a few mini games, 
in which the agent, for example, manipulates a \textit{Marine} to reach a beacon (\textit{MoveToBeacon})
or manipulates several \textit{Marines} to defeat several \textit{Roaches} (\textit{DefeatRoaches}), etc. 
However, for \emph{full} games, the current AI agents are still far
from achieving human professional level performance.

This paper investigates AI agents for full games, and for simplicity, we restrict our study to the following setting: 1vs1 Zerg-vs-Zerg on the AbyssalReef map. 
We develop two AI agents --- 
the AI agent~\AIRL~is based on deep reinforcement learning over flat actions and the AI agent~\AIHC~is based on rule controllers over hierarchical actions. 
Both \AIRL~and \AIHC~are able to defeat the built-in AI agents from level 1 to level 10 in a full game, 
noting that level 8, level 9, and level
10 are cheating agents with unfair advantages such as full vision on the whole map and resource harvest boosting.
It is also worth mentioning that according to some informal
discussions from the StarCraft II forum, level 10 built-in AI is
estimated to be Platinum to Diamond~\cite{scii-forum}, which are equivalent to top 50\% - 30\% human players in the ranking system of Battle.net Leagues~\cite{liquid}. 

The AI agent~\AIRL~is based on ``flat'' action modeling,
which employs a flat action structure and produces a number of discrete actions.
This design makes it immediately ready for any off-the-shelf RL
algorithm that takes discrete actions as input.
The AI agent~\AIHC~is based on ``deep'' action modeling,
which employs a manually specified action hierarchy.
Intuitively, ``deep'' modeling can better capture the action dependencies.
However, the training becomes more challenging since learning methods for
the complex \emph{hierarchical} RL will be needed. To avoid the
difficulty, we simply adopt a rule based controller for \AIHC~in this work.

To the best of our knowledge, this is the first public work to investigate the AI agents that can defeat the built-in AIs in a Starcraft II full game. The code will be open sourced~\cite{tstarbots}. We hope the proposed framework will be beneficial for future research in several possible ways:
1) Be a baseline for a hybrid system, in which more and more learning
modules can be included while rules are still utilized to express
logic that are hard to learn; 
2) Generate trajectories for imitation learning; and 
3) Be an opponent for self-play training.


\section{Related Work}
\label{sec:related-work}
The RTS game StarCraft I has been used as a platform for AI research
for many years. Please refer to~\cite{ontanon2013survey,churchill2016heuristic} for a review.
However, most research considers searching algorithms or multi-agent algorithms
that cannot be directly applied to the full game.
For example, many multi-agent reinforcement learning algorithms have been proposed to learn agents either independently~\cite{tan1993multi} or jointly~\cite{foerster2017counterfactual,usunier2016episodic,peng2017multiagent,sukhbaatar2016learning} with communications to perform collaborative tasks, 
where a StarCraft unit (e.g., a \textit{Marine}, a \textit{Zealot}, a \textit{Zergling}, etc.) is treated as an agent.
These methods can only handle mini-games, 
which should be viewed as snippets of a full game.

A vanilla A3C~\cite{mnih2016asynchronous} based agent is tried in SC2LE for full games~\cite{sc2le}, 
but the reported performance was relatively poor.
Recently, relational neural network is proposed for playing the SC2 game~\cite{zambaldi2018relational} using a player-level modeling. 
However, the studies were conducted on mini-games instead of full games.

Historically, some rule based decision systems have been successful in specific domains, 
e.g., MYCIN for medical diagnosis or DENTRAL for molecule discovery~\cite{russell2016artificial}.
The only way they can be improved is by manually adding knowledge,
which makes them lacking the ability to learn from data or from
interacting with  the environment.

Rule based AI-bot is popular in the video game industry.
However, the focus there is to develop tools for code reuse 
(e.g., Finite State Machine or Behavior Tree~\cite{millington2009artificial}),
not on how the rules can be combined with learning based methods.
There exists recent work that tries to perform reinforcement learning
or evolutionary algorithms over Behavior
Trees~\cite{marzinotto2014towards, perez2011evolving}. 
However, the observations are tabular, and are unrealistic for 
large scale games such as SC2. 

Our macro action based agent (Section~\ref{sec:rl-bot}) is similar to
that  of~\cite{tian2017elf}, 
where the authors adopted macro actions for a customized mini RTS game.
However, our macro action set is much larger. It encodes concrete rules for the execution,
and is therefore more realistic for SC2LE.

In Section~\ref{sec:rule-bot}, we describe our implementation of the hierarchical action based agent.
Our approach is inspired by the modular design of UAlbertaBot~\cite{churchill2016heuristic},
which is also widely adopted in the literature of StarCraft I.
In spirit, the hierarchical action set is similar to the FeUdal network~\cite{vezhnevets2017feudal},
but we do not pursue an end to end learning of the complete hierarchy.
We also allow each action to have its own observation and policy. 
This helps the system to rule out noisy information, as discussed in~\cite{van2017hybrid}.

\section{The Proposed TStarBot Agents}
\label{sec:method}
Among the multiple challenges of SC2, 
this work focuses on how to deal with the huge action space,
which, we argue, arises from the game's complex intrinsic structure.
Specifically, there are several aspects.

\textbf{Hierarchical nature.}
In RTS games, the long-horizon decision process requires complex hierarchical actions.
A human player often summarizes his or her thinking in several
abstraction levels including global strategies, local tactics, and micro executions.
If a learning algorithm is unaware of the higher ``abstraction
levels'' (i.e., the action hierarchy) and works directly on the the
huge number of basic atomic actions in full game play,
then it is inevitably difficult for RL training, especially due to
inefficient exploration.
For example, PySC2~\cite{sc2le} defines the action space over the low-level human user interface, involving hundreds of hot-keys and thousands of mouse-clicks over screen coordinates.
Following this setting, 
even the state-of-the-art RL algorithm can only achieve success in playing toy mini-games that has much shorter horizon than the full game~\cite{zambaldi2018relational}.
Although many papers have been devoted to automatically learning
meaningful hierarchies in a Markovian Decision Process~\cite{sutton1999between,bacon2017option,ghazanfari2017autonomous,vezhnevets2016strategic,vezhnevets2017feudal,frans2017meta}, 
none of these methods can work efficiently on environments as complex  as SC2. 
Therefore, it is a challenging task to utilize the hierarchical nature of the decision
process, and define a tractable decision space with manageable
exploration efforts.

\textbf{Hard-rules in SC2 are difficult to learn.}
Another challenge of learning-based agent design is the large number of  ``hard rules'' in RTS game. 
These hard rules may be considered as the ``laws of physics'' that
cannot be violated. 
They are easily interpreted by human players through in-game textual instructions, 
but are difficult for learning algorithms to discover using a pure
trial-and-error approach.
Consider a human player who starts to play StarCraft-II, 
he or she can easily learn from a textual tutorial to first select a \textit{drone} unit to build a \textit{RoachWarren} unit before selecting a \textit{larva} unit to produce a \textit{Roach} unit.
By doing so, he or she can quickly discover the following hard game rules:
\begin{itemize}
  \item \textit{RoachWarren is a prerequisite of producing a Roach.}
  \item \textit{RoachWarren is built by a Drone.}
  \item \textit{Roach is produced from a Larva.}
  \item \textit{Producing a Roach requires 75 minerals and 25 gas.}
  \item \textit{......}
\end{itemize}
In SC2 there are thousands of such dependencies,
constituting a technology dependency tree, abbreviated as \textit{TechTree} (See also Section~\ref{sec:pysc2}).
\textit{TechTree} serves as the most important prior knowledge that a human player should learn from textual tutorials or materials on the game interface, other than exploration through trial-and-error.
A learning algorithm unaware of the TechTree  must spend a huge amount of time to
learn the hard games rules,
which introduces extra difficulty, especially when the feedback signal is sparse and delayed (i.e., the \textit{win/loss} reward received at the end of each game).
Thus, in RTS games, it is important to think about how to design a
mechanism that can encode these hard game rules directly into the agent's prior knowledge, instead of relying on pure learning.

\textbf{Uneconomical learning for trivial decision factors.}
It is also worth noting that despite the tremendous decision space of SC2, not all decisions matter.
In other words,
a considerable amount of decisions are redundant in that they will have negligible effects on the game's final outcome.
For instance, when a human player wants to build a
\textit{RoachWarren} during the game, there are at least three
decision factors he or she has to consider:
\begin{itemize}
  \item Decision Factor 1: When to build it? (Non-trivial)
  \item Decision Factor 2: Which \textit{Drone} builds it? (Trivial)
  \item Decision Factor 3: Where to build it? (Trivial)
\end{itemize}
A proficient player would conclude that:
1) the first decision factor is a non-trivial one since when to build
a \textit{RoachWarren} will have a considerable impact on the entire game progress;
2) the second decision is trivial because any random \textit{Drone} can do the work with negligible difference of building efficiency; and
3) the third factor can also be taken as trivial as long as the target position is not too far away from the self-base and the \textit{geometry defense} is not considered.
Learning algorithms unaware of the factors may spend significant efforts
on filtering out trivial decisions.
For example, an accurate placement decision of ``where to build''
requires a selection among thousands of 2-D coordinates. 
It is thus uneconomical to invest too many learning resources for such trivial factors. 

To address these challenges, 
we propose to model the action structure with hand-tuned rules.
By doing so, 
the available actions are reduced to a tractable number of macro-actions, 
and the controller for the overall decision making system is easier to design.
Along this line of thought, we implemented two AI agents.
One agent adopts a reinforcement learning based controller over
pre-defined macro actions (Section \ref{sec:rl-bot}), 
while the other employs a macro-micro hierarchical action space with a
rule based controller (Section \ref{sec:rule-bot}). 
Execution of the actions relies on a per-unit-control interface of the
SC2 game, 
which is implemented in our PySC2 extension (Section \ref{sec:pysc2}).

\subsection{Our PySC2 Extension}
\label{sec:pysc2}
SC2LE~\cite{sc2le} is a platform jointly developed by DeepMind and Blizzard. The game core library provided by Blizzard exposes a raw interface and a feature map interface.
The DeepMind PySC2 environment further wraps the core library in Python and fully exposes the feature map interface.
The purpose is to closely mimic human controls (e.g., a mouse
click, or pressing some keyboard button), 
which introduces a huge number of actions due to the complexity of SC2. 
It thus poses difficulties for the underlying decision making system. 
Moreover, such a ``player-level'' modeling is inconvenient for
designing ``unit-level'' models, especially when multiple agents are considered.
In this work, we make additional efforts to expose unit level
controls.
Also, we encode the aforementioned building dependencies into a technology tree.

\textbf{Expose unit control.} In our PySC2 extension, we expose the raw interface of the SC2 core library,
which enables per unit observations and manipulations. 
At each game step, 
all units visible to the player (depending on whether fog-of-war is enabled) can be retrieved.
Each unit is fully described by a property list, 
including properties such as its position and health, etc.
Such a raw unit array is part of the observation returned to the agent.
Meanwhile, a per unit action is allowed to control each unit.
The agent can send raw action commands to interested individual units
(e.g., to move a unit to somewhere, or to ask a unit to attack another unit, etc.).
The definition of a unit and per-unit-actions can be found in the
protobuf from the SC2 core library.

\textbf{Encode the technology tree.} In Starcraft II, a player may need
particular units (or buildings or techs) as prerequisites for other
advanced units (or buildings or techs). 
Following UAlbertaBot~\cite{churchill2016heuristic}, 
we formalize these dependencies into
a technology tree, abbreviated as \textit{TechTree} in our PySC2 extension.
We have collected the complete \textit{TechTree} for \textit{Zerg},
which gives the cost, building time, building ability, builder, prerequisites for each \textit{Zerg} unit.

Besides the two additional functions described above, our PySC2 extension is fully compatible with the original Deepmind PySC2.

\subsection{\AIRL : A Macro Action Based Reinforcement Learning Agent}
\label{sec:rl-bot}

\begin{figure}[t]
\center
\includegraphics[width=0.9\linewidth]{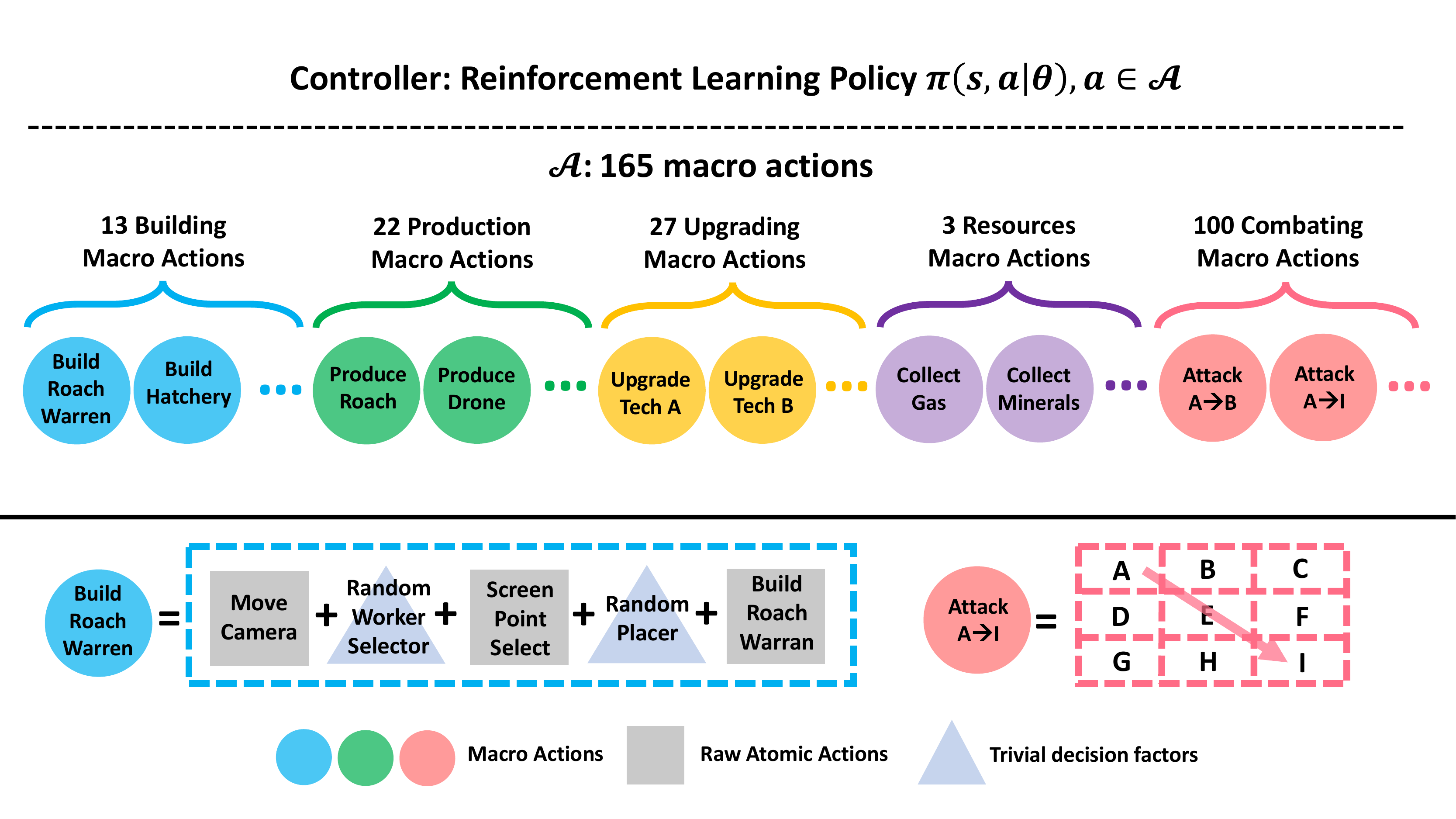}
\caption{\textbf{Overview of the agent based on macro action and reinforcement learning.}
At the top: a learnable controller over the macro actions exposed from the bottom;
At the bottom: a pool of 165 executable macro actions, which hard-code
prior knowledge of game rules (e.g. TechTree) and hide  the trivial
decision factors (e.g. building placement) and some execution details
from the top controller.
The figure also illustrates the definitions of two macro actions as examples: \textit{BuildRoachWarren} and \textit{ZoneAAttackZoneI}.}
\label{fig:rl-bot-overview}
\end{figure}

We illustrate in Figure~\ref{fig:rl-bot-overview} how the agent works. 
At the top, there is a single global controller, which will be
learned by RL and it makes decisions over macro actions that are exposed to it.
At the bottom, there is a pool of macro actions, 
which hard-codes prior knowledge of game rules
(e.g. \textit{TechTree}) and how actions are executed (e.g., which drone builds and where to build for a building action). 
Therefore it hides  trivial decision factors and executing details
from the top-level controller.

With this architecture, we relieve the underlying learning algorithm
from the heavy burden of directly handling a massive number of atomic
operations, 
while still preserving most of the key decision flexibilities of the full-game's macro strategies.
Moreover, such an agent can be equipped with basic knowledge of hard
game rules without learning. 
With such an abstraction of action space enriched with prior-knowledge, the agent can learn fast from scratch and beat the most difficult built-in bots within $1\sim2$ days of training over a single GPU. {More details are provided in the following subsections.}

\subsubsection{Macro Actions}
\label{sec:macro-action}
We designed 165 macro actions for the \textit{Zerg}-vs-\textit{Zerg} SC2 full-game, as summarized in Table~\ref{tab:macro_action_category} (please refer to Appendix-I for the full list).
As explained above, the purpose of the macro actions are two-fold:
\begin{enumerate}
  \item To encode the game's intrinsic rules that are difficult to
    learn using only the trial-and-error approach.
  \item To hide trivial decisions from the learning algorithm by hard-coded decision making.
\end{enumerate}
Each macro action executes a meaningful elementary task, e.g., to
build a certain building, to produce a certain unit, to upgrade a
certain technology, to harvest a certain resource,  to attack a certain place, etc..
Therefore it consists of a composition or a series of atomic operations. 
With such an abstraction in action space, learning a high-level
strategy for the full game becomes easier.
Some examples of macro actions are illustrated in Table~\ref{tab:macro_action_category}.

\begin{table}[t!]
  \centering
  \small
  \caption{\textbf{Summary of 165 macro actions}: their categories, examples and the hard-coded rules/knowledge.
  In the rightmost column, \textit{TechTree} has been explained in~\ref{sec:pysc2};
  \textit{RandUnit} refers to randomly selecting a subject unit;
  \textit{RandPlacer} refers to randomly selecting a valid placement coordinate.
  }
  \label{tab:macro_action_category}
  \resizebox{\textwidth}{15mm}{
  \begin{tabular}{|c|r|l|l|}
    \hline
    Action Category & \# & Examples & Hard-coded rules/knowledge\\
    \hline \hline
    Building & 13 & \textit{BuildHatchery},\textit{BuildExtractor} & TechTree, RandUnit, RandPlace \\
    \hline
    Production & 22 & \textit{ProduceDrone}, \textit{MorphLair} & TechTree, RandUnit \\
    \hline
    Tech Upgrading & 27 & \textit{UpgradeBurrow}, \textit{UpgradeWeapon} & TechTree, RandUnit\\
    \hline
    Resources Harvesting & 3 & \textit{CollectMinerals}, \textit{InjectLarvas} & RandUnit  \\
    \hline
    Combating & 100 & \textit{ZoneBAttackZoneD} & Micro Attack/Rally\\
    \hline
  \end{tabular}
  }
\end{table}

\textbf{Building Actions:}
Buildings are prerequisites for further unit production and tech upgrading in SC2.
The building category contains 13 macro actions, each of which builds a certain \textit{Zerg} building  when executed.
For example, the macro action \textit{BuildSpawningPool} builds a \textit{SpawningPool} unit with a series of atomic \textit{ui-actions}
\footnote{\textit{ui-actions} refers to the actions of the \textit{ui-control} interface in PySC2, resembling the human-player interface.
In fact, we use in this project the \textit{unit-control} interface (as described in Sec~\ref{sec:pysc2}) which simplifies the execution path by allowing agents to directly push action commands to each individual unit without having to first highlight-select a subject unit before issuing a command to it.}:
1) \textit{move\_camera} to base,
2) \textit{screen\_point\_select} a drone (the subject unit),
3) \textit{build\_spawningpool} somewhere in the screen.
The series of atomic operations have two internal decisions to make:
1) which \textit{Drone} is used to build it? and
2) where to build it?
Since these two decisions usually have little impact on the entire
game process, we employ random and rule-based decision-makers,
namely, a random \textit{Drone} selector and a random spatial placer.
The random spatial placer has to encode the basic placement rules
like: \textit{Zerg} buildings can only be placed on the \textit{Creep}
zone; \textit{Hatchery} has to be located near minerals for fair
harvesting efficiency. 
In addition, the \textit{TechTree} rules such as ``only \textit{Drone} can build \textit{SpawningPool}'' are also encoded in this macro action.

\textbf{Production \& Tech Upgrading Actions:}
Unit production and tech upgrade largely shape the economy and technology development in the game.
The production category contains 22 macro actions and the tech upgrade
contains 27. Each of the macro actions either produces a certain type of units or upgrades a certain technology.
These macro actions are hard-coded, similar to the building actions described above, 
except that they do not need a spatial placer.

\textbf{Resource Harvesting Actions:}
\textit{Minerals}, \textit{Gas}, and \textit{Larvas} (\textit{Zerg}-race only) are the three key resources in SC2 games.
Their storage and collection speed can greatly affect the economy growth.
We designed 3 corresponding macro actions: \textit{CollectMinerals}, \textit{CollectGas} and \textit{InjectLarvas}.
\textit{CollectMinerals} and \textit{CollectGas} assign a certain number of random workers (i.e. \textit{Drone} in \textit{Zerg}-race) to mineral shards or gas extractors, so that with these two macro actions, the workers can be re-allocated to different tasks, altering the mineral and gas storage (or their ratio of storage) to meet certain needs.
\textit{InjectLarvas} simply orders all the idle queens to inject
\textit{Larvas}, with the effect of speeding up the unit production process.

\textbf{Combat Actions:}
Combat action design is the most important aspect of SC2 AI agents,
directly affecting the game's outcome. We have carefully
designed macro actions to handle the following 
aspects of combat strategies.
\begin{itemize}
  \item Attack timing: e.g., rush, early harass,  the best attack timing windows.
  \item Attack routes: e.g., walk around narrow slopes which might
    constrain the attack firepower.
  \item Rally positions: e.g., rally before attack in order to concentrate fire (note that various units might have different moving speed).
\end{itemize}

We represent these combat strategies by region-wise macro actions,
which are defined as follows (also see Figure~\ref{fig:rl-bot-overview}).
We first divide the entire world map into nine combat zones (named
\textit{Zone-A} to \textit{Zone-I}), and an additional \textit{Zone-J}
for the entire world itself, resulting in 10 zones in total.
Based on the zones, 100 ($=10 \times 10$) macro actions are defined,
with each macro action executing rules such as:
``combat units in \textit{Zone-X} start to attack \textit{Zone-Y} if there are enemies there, otherwise, rally to \textit{Zone-Y} and wait''.
For the micro attack tactics inside each macro action, we simply hard-code the ``hit-and-run'' rule for each combat unit,
i.e., 
the unit fires at the closest enemy and runs away upon low health.
We leave the investigation of more sophisticated multi-agent learning of micro tactics to future work.

With the composition of these macro actions, a wide range of diverse macro combat strategies can be represented.
For example, the selection of attack routes can be represented by a series of region-wise rally macro actions. With this definition, we avoid the difficulty of complex multi-agent learning.

\textbf{Available Macro Action List.}
Not every pre-defined macro action described above is available at any time step.
For example, there are constrains in the \textit{TechTree}  indicating
some units/techs can only be built/produced/upgraded under certain
conditions: e.g., having enough storage of minerals/gas/food or the existence of certain prerequisite unit/tech.
The corresponding macro action should be ``do nothing'' when these conditions are not satisfied. 
We maintain a list of such available macro actions at each time step,
encoding the \textit{TechTree} knowledge.
This list of available actions masks the invalid actions at each time step.
The list can also be used as features for machine learning.

\subsubsection{Observations and Rewards}
\label{sec:obs_reward}

The observations are represented as a set of spatial 2-D feature maps
and a set of non-spatial scalar features, extracted from the per-unit
information provided by the SC2 game core, which is exposed in our PySC2 extension (see Section~\ref{sec:pysc2}).

\textbf{Spatial Feature Maps.}
Extracted feature maps are of size $N \times N$, where $N$ is smaller
than the screen dimension. Each pixel of a feature map
corresponds to a small region in the entire world map, 
representing a certain statistical quantity such as the
unit count of a certain type in the region.
These quantities include the counts of commonly-used unit types both
for the player and for the opponent, and unit counts
with certain attributes such as ``can-attack-ground'' and ``can-attack-air''.

\textbf{Non-spatial Features.}
Scalar features include the amount of gas and minerals collected, the amount of food left, the counts of each unit types, etc.
They also include recently-taken actions to keep track of the past information.

\textbf{Rewards.}
We use a ternary valued reward function: 1 (win) / 0 (tie) / -1 (loss)
received at the end of a game. The reward is always zero during the game.
Although the reward signal is quite sparse, and has a long
time-horizon delay, it nevertheless works with the macro action
structure presented in this work.

\subsubsection{Learning Algorithms and Neural Network Architectures}

Based on the macro actions and observations defined above, the problem
becomes a sequential decision process,
where the macro action space is of a tractable size and the number of
macro action
steps in a game is shortened from the number of original atomic actions.

At time step $t$, an agent receives an observation $s_t \in \mathcal{S}$ from the game environment,
and chooses a macro action $a_t \in \mathcal{A}$ according to its policy $\pi(a_t | s_t)$, a conditional probability distribution over $\mathcal{A}$, 
where $\mathcal{A}$ indicates the set of macro actions defined in Section~\ref{sec:macro-action}.
The selected macro action $a_t$ is then translated into a sequence of
atomic actions that are acceptable to the game-core by using the corresponding hand-tuned rules.
After the atomic actions are taken, a reward~\footnote{
The reward is accumulated within the macro action's execution time, if
the macro action lasts for multiple time-steps.} $r_t$ and the next
step observation $s_{t+1}$ are received by the agent. This loop goes on until the end of a game.

Our goal is to learn an optimal policy $\pi^*(a_t | s_t)$ for the agent to maximize its expected cumulative rewards over all future steps.
When we directly use the reward function defined in
Section~\ref{sec:obs_reward} without additional reward shaping, the
optimization target is equivalent (when reward discount is ignored) to
maximizing the agent's probability of winning.
We train our {\AIRL} agent to learn such a policy from scratch by playing against built-in AIs
with off-the-shelf reinforcement learning algorithms (e.g.
Dueling-DDQN~\cite{mnih2015human,van2016deep,wang2016dueling} and
PPO~\cite{schulman2017proximal}), together with a distributed rollout infrastructure. Details are presented below.

\textbf{Dueling Double Deep Q-learning (DDQN).}
Deep Q Network~\cite{mnih2015human} first learns a parameterized estimation $\hat{Q}(s,a|\theta)$ of the optimal state-action value function (Q-function) $Q^*_{\theta}(s_t, a_t)=\max_{\pi}Q^{\pi}(s_t, a_t)$,
where $Q^{\pi}(s_t, a_t)=\mathbb{E}_{\pi}[\sum_{i=t,...,T} \gamma^{i-t} r_i]$ is the expected cumulative future rewards under policy $\pi$.
The optimal policy can be easily induced from the estimated optimal
Q-function: $\pi(a_t|s_t)=1.0$ if $a_t = \arg \max_{a \in \mathcal{A}}
\hat{Q}(s_t, a|\theta)$, and $\pi(a_t|s_t)=0$ otherwise.
Techniques such as replay memory~\cite{mnih2015human}, target
network~\cite{mnih2015human}, double networks~\cite{van2016deep} and
dueling architecture~\cite{wang2016dueling} are leveraged to reduce
sample correlation, maximization bias, update target inconsistency and
update target variance. These techniques improve learning stability and sample efficiency.
Due to the sparsity and long-delay of the rewards, we use a Mixture of Monte-Carlo (MMC)~\cite{bellemare2016unifying} return with the boostrapped Q-learning return as the Q update target,
which accelerates the reward propagation and stabilizes the training.

\textbf{Proximal Policy Optimization (PPO).}
We also conducted experiments by directly learning a parametric form of stochastic policy $\pi(s_t, a_t|\theta)$ with Proximal Policy Optimization (PPO)~\cite{schulman2017proximal}. PPO is a sample efficient policy gradient method, leveraging policy ratio trust region clipping to avoid the complex conjugate gradient optimization required to solve the KL-divergence constrained Conservative Policy Iteration problem in TRPO~\cite{schulman2015trust} . We used a truncated version of generalized advantage estimation~\cite{schulman2015high} to trade-off the bias and variance of the advantage estimation. The available action list described in Section~\ref{sec:macro-action} is used to mask out unavailable actions and renormalizes the probability distributions over actions at each step.

\textbf{Neural Network Architecture.}
We adopt multi-layer perception neural networks to parameterize the state-action value function, state value function and the policy function.
While more complex network architectures could be considered (e.g.,
convolutional layers that extracts spatial features, or recurrent
layers that compensates the partial observation), we will leave them
to future work.

\textbf{Distributed Rollout Infrastructure.}
The SC2 game core is CPU-intensive and slow for the rollout,
leading to a bottleneck during the RL training.
To alleviate the issue,
we build a distributed rollout infrastructure,
where a cluster of CPU machines (called actors)
are utilized to perform the rollout processes in parallel. 
The rollout experiences, cached in the replay memory of each actor, are randomly sampled and periodically sent to a GPU-based machine (called learner).
We currently take 1920 parallel actors (with 3840 CPUs across 80 machines) to generate the replay transitions, at the speed of about 16,000 frames per second.
This significantly reduces the training time (from weeks to days), and also improves the learning stability due to the increased diversity of the explored trajectories.

\subsection{\AIHC : A Hierarchical Macro-Micro Action Based Agent}
\label{sec:rule-bot}
The macro action based agent described in Section~\ref{sec:rl-bot}
has some limitations.
Although the macro actions can be grouped according to functionality, 
a single controller has to work over all action groups,
where the actions of different groups are mutually exclusive at each decision step.
Also, when predicting what action to take,
the controller takes a common observation that is unaware of the action group.
This amounts to unnecessary difficulties for training the controller,
as undesired information may kick in for both observations and actions.
Moreover,
the macro action does not have any control over individual units (i.e., per-unit-control), 
which is inflexible when we want to adopt multi-agent style methodology.

\begin{figure}[t]
\center
\includegraphics[width=0.9\linewidth]{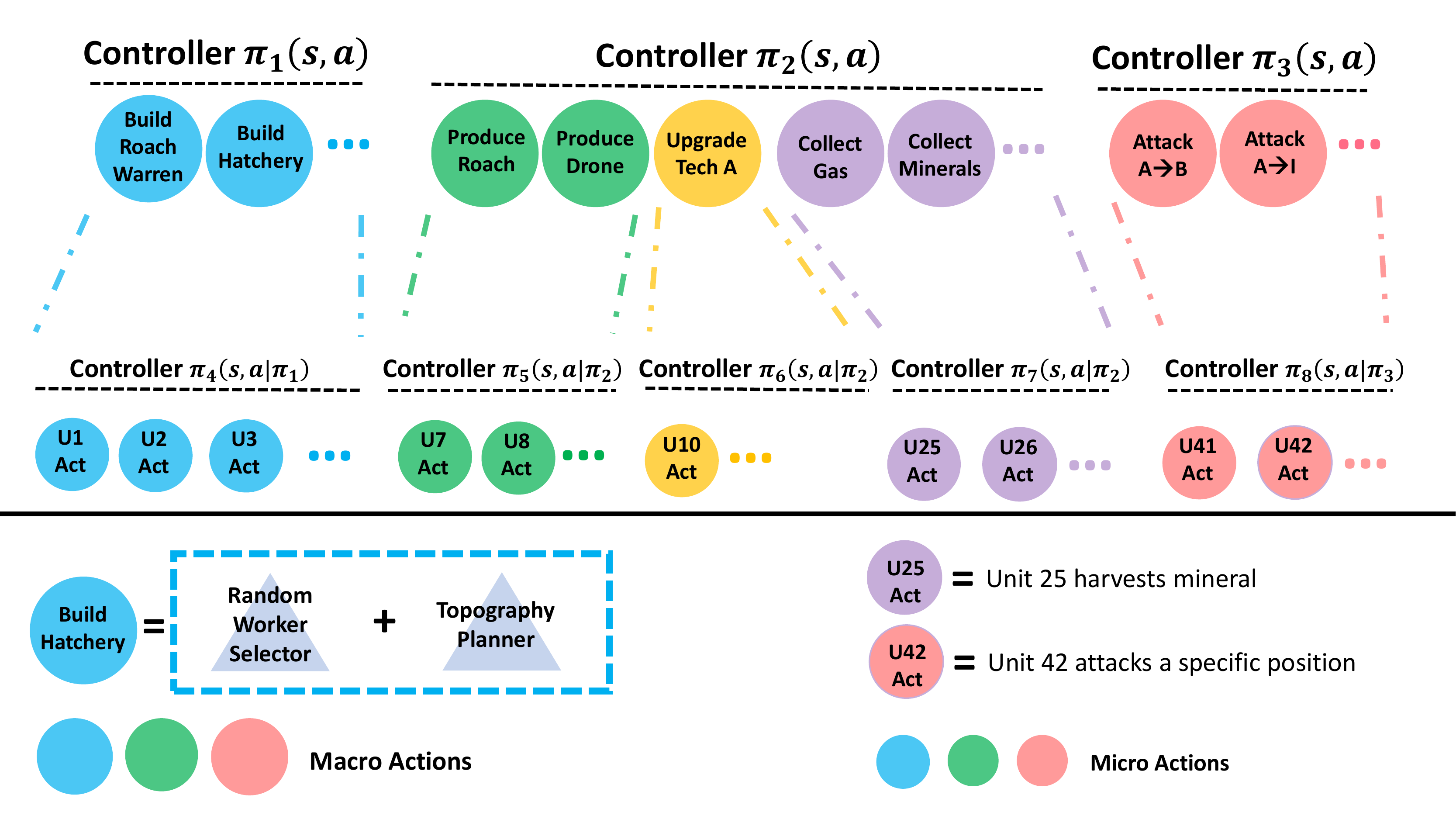}
\caption{\textbf{Overview of the macro-micro hierarchical actions.}
See the main text for explanations.
}
\label{fig:mm-overview}
\end{figure}

For improved flexibility, we have created a different set of actions, 
as in Figure~\ref{fig:mm-overview}.
We employ both macro actions and micro actions,
organized in a two-tier structure.
The upper tier corresponds to macro actions, 
which represent high-level strategies/tactics such as ``build RoachWarren near our main base'' or ``squad one attacks enemy base'';
while the lower tier corresponds to micro actions,
which correspond to low-level controls over each unit 
such as ``unit 25 builds RoachWarren at a specific position'' or ``unit 42 attacks to a specific position''.
The entire action set is divided into groups both horizontally and vertically.
Each action group is assigned a separate controller that can only see
the local actions and the local observations that are relevant to the actions therein. 
At each time step, the controllers at the same tier can take simultaneous actions,
while a downstream controller has to be conditioned on its upstream controller.   

There are two advantages of this hierarchical structure.
1) Each controller has its own observation/action space so that 
irrelevant information can be filtered out more easily;
this is also adopted and discussed in~\cite{van2017hybrid} when modeling the sub-task Q head.
2) The hierarchy captures the game's action structure better, with
simultaneous actions from different controllers.

Although ideally
the controllers should be trained with RL either separately or jointly,
in this work we simply employed expert rules, with the
intention of validating the proposed hierarchical action set approach.

\begin{figure}[h]
\center
\includegraphics[width=0.9\linewidth]{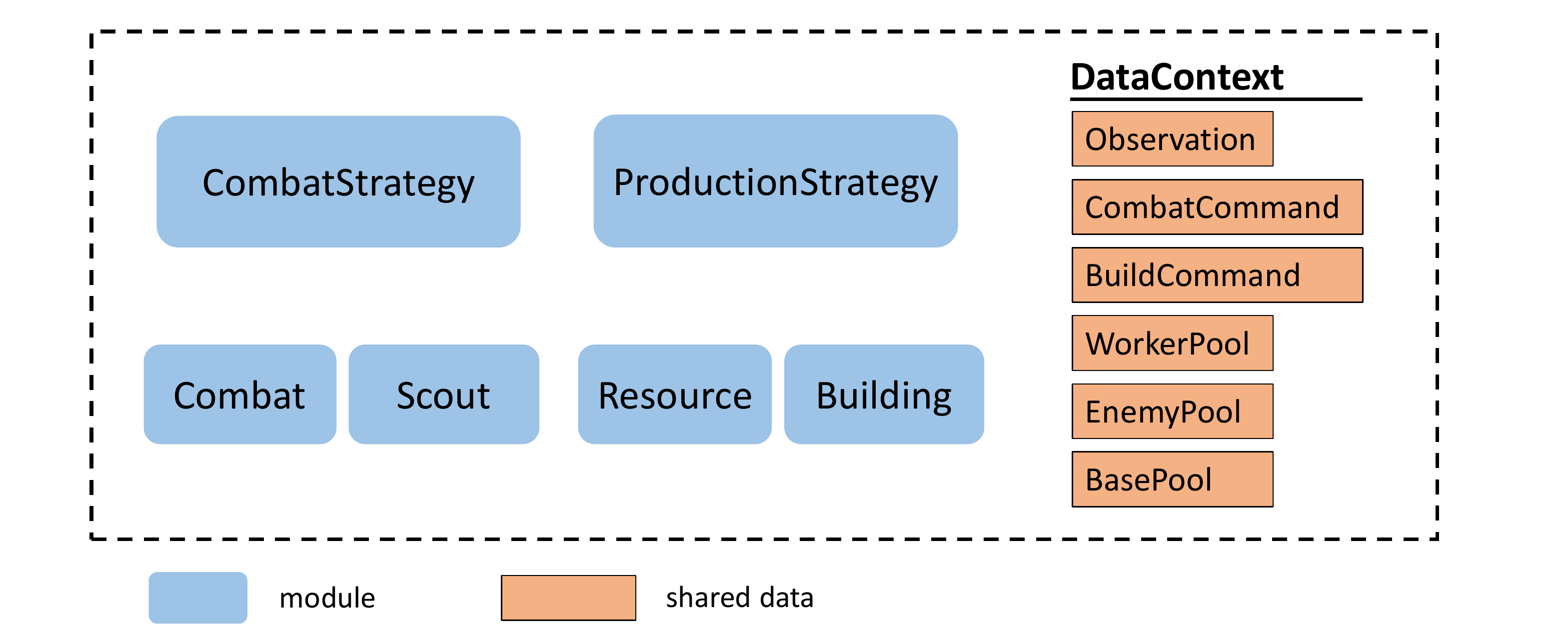}
\caption{\textbf{Module diagram for the agent based on the Macro-Micro Hierarchical Action.} See the main text for explanations.} 
\label{fig:rule-bot-overview}
\end{figure}

As in Figure \ref{fig:rule-bot-overview}, each controller represents a module,
organized in a way similar to UAlbertaBot.
The first tier modules (CombatStrategy, ProductionStrategy) only issue high-level commands (macro actions),
while the second tier modules (Combat, Scout, Resource and Building) issue low-level commands (micro actions).
All these modules are embedded into a DataContext, where each module can communicate with others by sending/receiving messages and sharing customized data structures. 
Crucially, 
the game-play observation exposed by PySC2 is placed in the DataContext and henceforth visible to every module. 
This way,
each module can extract  local observation relevant to its own action
set from a common observation.
In the following we describe the modules in greater details. 

\subsubsection{Data Context}
The DataContext module serves as a ``black board'' where the modules exchange information.
What contained in the DataContext fall into the following categories.
\begin{enumerate}
\item Observation. The feature maps provided by PySC2, as well as the
  unit data structure of all active units at the current time step, are exposed.
\item Pool. 
A pool is an array for a specific type of units,
with associated properties/methods for the easy access of caller module.
For example, the WorkerPool is the array of all \textit{Zerg Drones}. 
As another example, the BasePool is the array of all \textit{Zerg} bases, with
each item in the pool being a BaseInstance. 
The BaseInstance is a customized data structure that records the Base (can be \textit{Hatchery/Hive/Lair}), 
the associated \textit{Drones}, \textit{Minerals} and \textit{Extractors} within a fixed range of the Base, 
and a local coordinate system given by the geometrical layout of the minerals and the base. 
\item Command Queue. 
High level commands are stored in a queue visible to all (lower-tier) modules.
For instance, the commands issued by the ProductionStrategy module are pushed in BuildCommand.
A high level command may be ``update a particular technology'' or ``harvest more minerals currently''.
The lower-tier module (respectively Building and Resource in this
case) will pull from the queue a command it recognizes and execute it by
taking the corresponding rules to produce actions acceptable by the game core.
\end{enumerate}

At each time step, the DataContext will update the Observations and various Pools,
while the Command Queues will be modified or accessed by other modules.

\subsubsection{Combat Strategy}
The combat strategy module makes high-level decisions so that the
agent can combat enemies in different ways. 
The module manipulates all the combat units\footnote{\small Currently, the combat units do not include \textit{Drones} and \textit{Overlords}.} by organizing them into squads and armies. 
Each squad, which may contain one or multiple combat units, is
expected to execute a specific task, such as harassing an enemy base, cleaning the rock in the map, etc. 
Commonly, a small group of combat units with the same unit type is organized into a squad. 
An army contains multiple squads with a high-level strategic objective, e.g., attacking enemy, defending base, etc., and then specific commands are sent to each squad in the army. 
Each command, coupled with a squad-command pair, is then pushed into a combat strategy command queue (maintained in data context), which will be received and executed by the combat module.

Our implementation includes five high-level combat strategies: 
\begin{itemize}
	\item Rush: Once a squad of a small number of combat units has built up, launch attack and keep sending squads to attack the enemy base.
	\item Economy First: Collect minerals and
          gases first, and then launch attack after a large number of squads have been accumulated.
	\item Timing Attack: Build up a strong army of \textit{Roach} and \textit{Hydralisk} squads as quickly as possible, and initiate a strong attack.
	\item Reform: Sort enemy bases and let the army attack the
          closest enemy base with high priority. When approaching the
          target enemy base, stop the leading squads and let them wait
          for other squads to gather. Then, launch attack.
	\item Harass: Set the combat strategy for the ground combat
          units as `Reform'. Build up 2-3 squads of \textit{Mutalisk}
          and assign a target enemy base to each of them. 
          Then, let the \textit{Mutalisk} detour and harass the \textit{Drones} of the target enemy base.
\end{itemize}

\subsubsection{Combat}
The combat module fetches commands from the command queue and execute a specific action for each unit. 
It uses unit-level manipulation to effectively let each unit fight against the enemy. 
The combat module implements some basic human-like micro-management tactics, such as hit-and-run, cover-attack, etc., which can be deployed to all combat unit types. 
Specifically, an additional micro-management manger for each specific combat unit is implemented by taking full use of the unit-type-specific skills. 
For example, the \textit{Roach} micro-manager enables roaches to burrow down and run away from enemy to recover when they are weak; 
\textit{Mutalisks} are coded to stealthily reach the enemy base and harass enemy's economy; 
\textit{Lurkers} use carefully designed hit-and-run tactics in combination with burrowing down and up; 
and \textit{Queens} can provide additional \textit{Larvas} and cure weak allies, etc.
These micro-managements are organized into hierarchies and each part can be conveniently replaced with RL models.

\subsubsection{Production Strategy}
\label{sec:product}
The production strategy module manages the building/unit production, tech upgrading and resource harvesting. 
The module controls the production of units and buildings by sending production instructions to each BaseInstance.
The tech upgrading instructions and other specific instructions, 
such as \textit{Zerg's Morph}, are pushed precisely to the target unit.
Then the Building module will implement all of the above production instructions.
The resource harvesting command are highly abstract that the production strategy only 
needs to determine what is prioritized, gas or mineral, according to the mineral/gas storage ratio.
The Resource module will then re-allocate workers to each BaseInstance based on the priority instruction.

In the module, we maintain a building order queue for short-term production planning.
Most time, the manager will follow the order to produce items (units, buildings or techs)
as long as there are enough resources and the prerequisites are satisfied.
In some special cases (e.g., expanding a new base) or in emergency situations (e.g., find cloaked enemy units),
a more prioritized item can be put in front of the queue, or we can
even clear the entire queue when a new goal has been set.
When the queue is empty (including at the beginning of the game),
a new short-term goal should be set immediately.

When executing the actual production at each time step,
the prerequisites and resource requirement of the current item will be checked according to the TechTree.
The prerequisites of advanced items will be added into the queue automatically if required,
and the current time step will be skipped if the resource requirement is not satisfied.
Moreover, when the current item is ready to be produced,
a BaseInstance will be selected, informed with the item type, and assigned to perform the concrete production.

By using different opening order and goal planning functions,
we have defined two different production strategies for \textit{Zerg} as follows.
\begin{itemize}
	\item RUSH: “Roach rush”. It produces roaches at the beginning, and upgrades tech \textit{BURROW} and 
	\textit{TUNNELINGCLAWS} to give \textit{Roaches} the ability to burrow and move while burrowing,
	and to increase the health regeneration rate while burrowing.
	The strategy continuously produces Roaches and \textit{Hydralisks}.	 
	\item DEF\_AND\_ADV: "Defend and Advanced Armies". This strategy produces many \textit{SPINECRAWLERs}
	at the second base to defend, and then gradually produces advanced armies.
	Almost all types of combat units are included and the final
        ratio among the types is restricted according to a predefined dictionary.
\end{itemize}

\subsubsection{Building}

The building module receives and executes high level commands issued by the Production Strategy, as described in Section~\ref{sec:product}.
The ``unary'' commands (i.e., let some unit act by itself)
are straightforward to execute. 
Some ``binary'' commands require more explanations.

The command ``Expand'' will drag a drone from the specified base, 
send it to the specified ``resource area'', and start morphing a Hatchery, 
whose global coordinate is pre-calculated by a heuristic method when
the map information is obtained for the first time.  

The command ``Building'' will drag a drone, morph it into the specified building at some position, 
whose coordinate is decided by a dedicated sub module, called Placer. 
In our implementation, we adopt a hybrid method for building placement,
i.e., some of the core buildings are placed in predefined positions, 
while others are placed randomly.
Both of these two types of positions are in the BaseInstance local coordinate system, 
and will be translated into the global coordinate system when they are
converted into game core acceptable actions. 
Specifically, all tech upgrading related buildings and the first six \textit{SpineCrawlers} are pre-defined.
Note that the layout of the six \textit{SpineCrawlers} placement is critical (e.g., whether they are in diamond formation or in rectangular formation), 
affecting the quality of the defense and whether we can survive an early rush of the opponent player.  
We have tried several arrangements and decided on the diamond formation.
The other buildings, including additional \textit{SpineCrawlers}, will be placed randomly,
where a uniformly random coordinate is generated repeatedly until it passes all validity checking (e.g., whether it is on \textit{Zerg} creep, whether it overlaps with other buildings, etc.). 

\subsubsection{Resource}
The resource module is to harvest minerals and gases by sending drones to either mineral shards or extractors. 
At each time step, this module needs to know whether the current working mode is ``mineral first'' or ``gas first'', 
which is a high level command, called ``resource type priority'' issued by the Production Strategy module.
The goal of this module is to maximize the resource collecting speed,
which can be a complex control problem.
In our implementation, we adopt several rules to achieve this goal, 
which turns out to be simple yet effective.
The underlying idea is to let every drone work and avoid any drone being idle.
Specifically, we let the following rules to be executed sequentially at each time step.

\begin{enumerate}
\item Intra-base rules. 
At each time step, the local drones associated with a BaseInstance will be rebalanced to harvest more minerals or more gases, depending on the ``resource type priority'' command.
Note that for each base and extractor the SC2 game core maintains two useful variables ``ideal harvesters number'', 
which is the suggested maximum number of drones working on it, 
and ``assigned harvesters number'', 
which is the actual number of drones working on it.
Using these two variables, it is easy to decide whether the local
working drones for minerals and gases are under-filled or over-filled.
\item
Inter-base rules.
When a new branch base is about to finish, 
drag 3 drones from other bases into the new base.
This improves the resource collecting efficiency by saving some waiting time.
We find this trick to be important, especially when expanding the first branch base.
\item
Global rules.
It scans for possible idle workers. 
Each idle worker is sent to the nearest base to harvest either mineral or gas, depending on the current working mode ``resource type priority''.
Note that when minerals or extractors are exhausted and all local drones working on them become idle, 
the rules also ensure that these idle workers are sent to nearby bases. 
\end{enumerate}

\subsubsection{Scout}
The Scout module tries to find out as many enemy units as possible.
With the fog-of-war mode enabled, each unit has only a very confined view.
Consequently, many enemy units are invisible, 
unless the player's own units can approach them and see them via {\em scouting}.

In our implementation, we send \textit{Zerg Drones} or \textit{Overlords} to detect enemy units and store the discovered units in EnemyPool,
from which we can infer high level information, 
such as the location of the enemy main base or branch base,
current buildings of the enemy, etc.
This kind of information can be further used to infer enemy's strategy, 
useful for the CombatStrategy or ProductionStrategy to make counter-strategy accordingly.

We define the following scout tasks.
\begin{enumerate}
\item
Explore Task. 
Whenever there is a new \textit{Overlord}, we send it to a mineral zone.
This action tries to look at the territory of the enemy in order to infer its economy.
When attacked, the \textit{Overlord} will retreat; otherwise it just stays at the target position. 
\item
Forced Task. 
We send a Drone to the enemy's first branch base.
By doing so, we can find out useful informations (e.g., whether a lot of
enemy \textit{Zerglings} have rallied, which happens when the enemy is about to perform a RUSH strategy at the early stage of the game play). 
\end{enumerate}
The activation of each task depends on the game's progress and time steps.

\section{Experiment}
\label{sec:exp}
Experimental results are reported for the two agents described in Section~\ref{sec:rl-bot} and Section~\ref{sec:rule-bot}, respectively.
We have tested the agent in a 1v1 \textit{Zerg}-vs-\textit{Zerg} full game.
Specifically, the agent plays against the built-in AIs ranging from level 1 (the easiest) to level 10 (the hardest).
The map we use is AbyssalReef\footnote{This map is an official map widely used in world class matches.},
on which a vanilla A3C agent over the original PySC2
observations/actions was reported~\cite{sc2le}, although it performed poorly when playing against built-in AIs in a Terran-vs-Terran full game.

\begin{figure}[t]
\center
\includegraphics[width=1.0\linewidth]{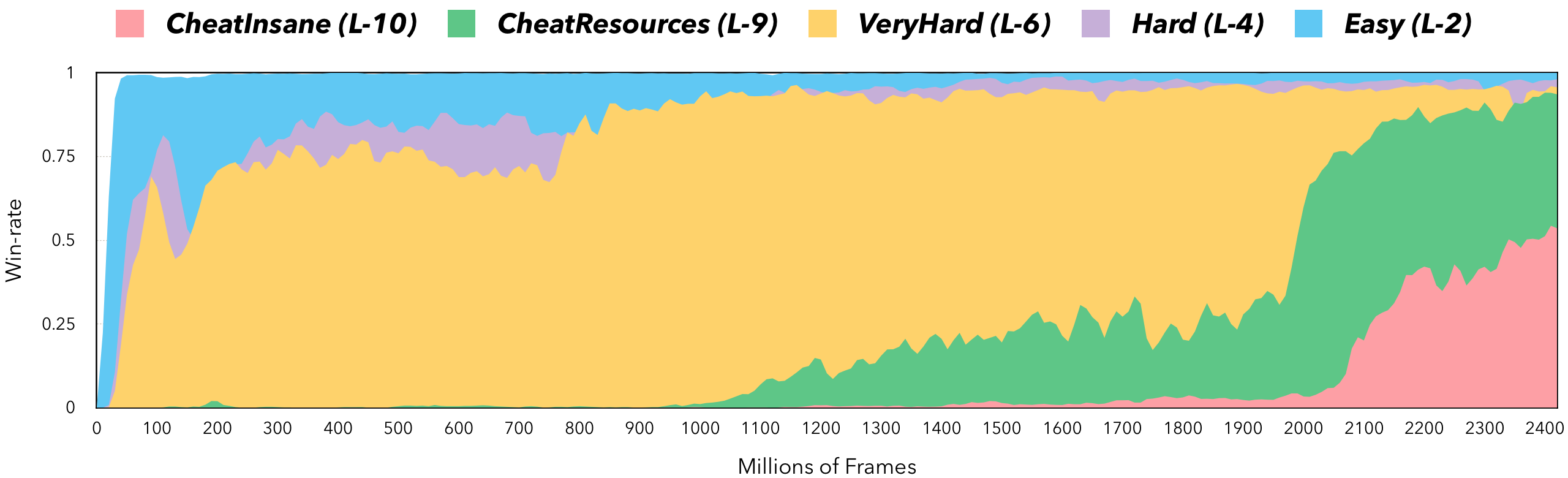}
\caption{Learning curves of {\AIRL} with PPO algorithm. Note that {\AIRL} - PPO starts to defeat (at least 75\% win-rate) \textit{Easy} (Level-2) built-in AI at about 30M frames, \textit{Hard} (Level-4) at about 250M frames, \textit{VeryHard} (Level-6) at about 800M frames, \textit{CheatResources} (Level-9) at about 2000M, and \textit{CheatInsane} (Level-10) at about 3500M frames.}
\label{fig:ppo_learning_curve}
\end{figure}

\subsection{\AIRL}

The proposed macro-action-based agent \AIRL (Section~\ref{sec:rl-bot}) is trained by playing against a mixture of built-in AIs in various difficulty levels: 
for each rollout episode, a difficulty level is sampled uniformly at random from level-1, 2, 4, 6, 9, 10 for the opponent built-in AI.
We restrict {\AIRL} to take one macro action every 32 frames (i.e. about every 2 seconds), 
which shortens the time horizon to about $300 \sim 1200$ steps per game and
reduces {\AIRL}'s APM (Actions Per Minute) to about $400 \sim 800$,
which is more comparable with that of human players. 
In these preliminary experiments, we only use non-spatial features
together with a simple MLP neural network. 
Also, in order to accelerate learning, we prune the combat macro
actions and we only use \textit{ZoneJ-Attack-ZoneJ}, \textit{ZoneI-Attack-ZoneD}, \textit{ZoneD-Attack-ZoneA}.

Table~\ref{tab:win_rate} reports the win rates of {\AIRL} agent against built-in AI ranging from level 1 to level 10.
\begin{table}[]
  \centering
  \caption{\textbf{Win-rate (in \%) of {\AIRL} and {\AIHC} agents, against built-in AIs of various difficulty levels}. For {\AIRL}, results of DDQN, PPO, and a random policy are reported. Each win-rate is obtained by taking the mean of 200 games with different random seeds, with Fog-of-war enabled.
  }
  \vspace{0.1in}
  \label{tab:win_rate}
  \resizebox{\textwidth}{15mm}{
  \begin{tabular}{l|c|c|c|c|c|c|c|c|c|c|c}
    \hline
    \multicolumn{2}{l|}{Difficulty Level IDs} & L-1 & L-2 & L-3 & L-4 & L-5 & L-6 & L-7 & L-8 & L-9 & L-10 \\
    \hline
    \multicolumn{2}{l|}{\tabincell{l}{Difficulty Level\\Descriptions}} & \tabincell{c}{Very\\Easy} & \tabincell{c}{Easy} & \tabincell{c}{Medium} & Hard & Harder & \tabincell{c}{Very\\Hard}  & Elite & \tabincell{c}{Cheat\\Vision} & \tabincell{c}{Cheat\\Resources} & \tabincell{c}{Cheat\\Insane} \\
    \hline \hline
    \multirow{3}{*}{{\AIRL}}
    & RAND & 13.3 & 0.0 & 0.0 & 0.0 & 0.0 & 0.0 & 0.0 & 0.0 & 0.0 & 0.0 \\
    \cline{2-12}
    & DDQN & 100.0 & 100.0 & 100.0 & 98.3 & 95.0 & 98.3 & 97.0 & 99.0 & 95.8 & 71.8 \\
    \cline{2-12}
    & PPO & 100.0 & 100.0 & 100.0 & 100.0 & 99.0 & 99.0 & 90.0 & 99.0 & 97.0 & 81.0 \\
    \hline \hline
    \multicolumn{2}{l|}{{\AIHC}} & 100.0 & 100.0 & 100.0 & 100.0 & 100.0 & 99.0 & 99.0 & 100.0 & 98.0 & 90.0 \\
    \hline
  \end{tabular}
  }
\end{table}
Each reported win-rate is obtained by taking the mean of 200 games with different random seeds,
where a tie is counted as 0.5 when calculating the win-rate.
After about $1\sim2$ days of training with a single GPU and 3840 CPUs, the reinforcement learning agent (both DDQN and PPO) can win more than 90\% of games against all built-in bots from level-1 to level-9, and more than 70\% against level-10.
The training and evaluation are both carried out with \textit{Fog-of-war} enabled (no cheating).

\begin{figure}[t!]
\centering
\hspace*{-0.2in}
\includegraphics[width=0.33\linewidth]{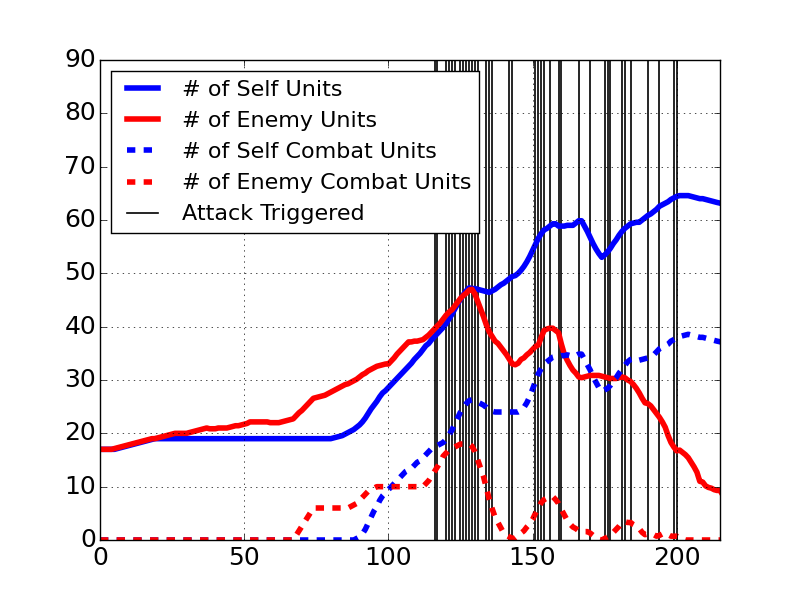}
\hspace*{-0.2in}
\includegraphics[width=0.33\linewidth]{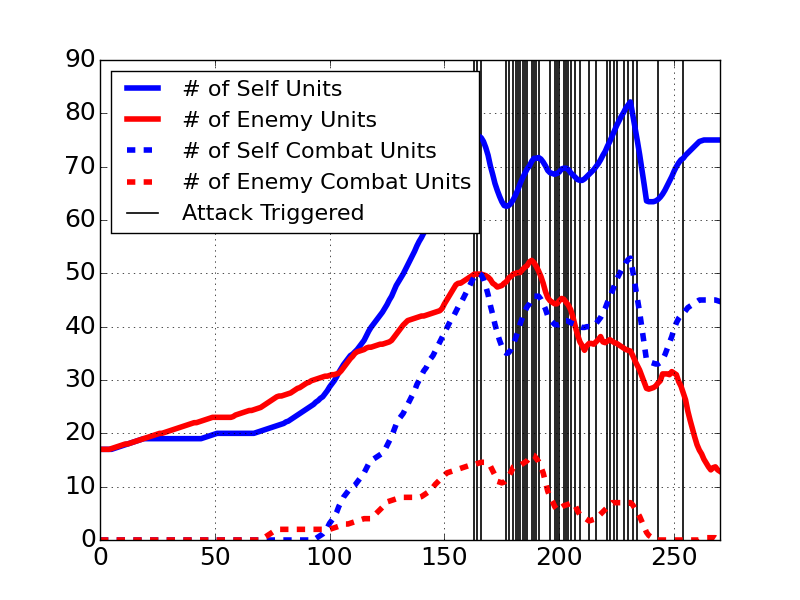}
\hspace*{-0.2in}
\includegraphics[width=0.33\linewidth]{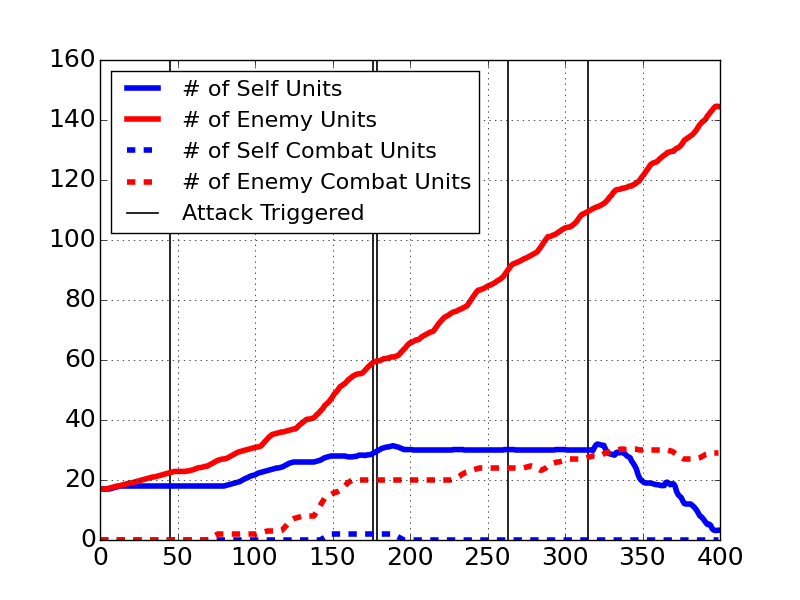}
\hspace*{-0.2in}
\caption{\textbf{The learned strategies about combat timing: \textit{Rush} and \textit{EconomyFirst}, for the {\AIRL} agent.}
In each figure we plot several in-game statistics: 
self units count (blue solid curves), enemy units count (red solid curves), 
self combat-units count (blue dashed curves), 
enemy combat-unit count(red dashed curves), 
and combat timing (black vertical lines).
The left and middle figures correspond to the learned RL policy,
while the right figure corresponds to a random policy.
The timing showed in the left figure resembles a human strategy called \textit{Rush}, which launches attacks as soon as possible, even if there are only a small number of combat units available;
The middle figure illustrates an \textit{EconomyFirst} strategy, which launches the first attack only after having assembled a strong enough army.}
\label{fig:rl-visualize}
\end{figure}

Figure~\ref{fig:ppo_learning_curve} shows the learning progress of {\AIRL} using the PPO algorithm. 
The curves show how the win-rate increases with the increased number
of frames being seen during training.
Each curve corresponds to a built-in AI at a certain difficulty level.
Note that {\AIRL}  starts to defeat (at least 75\% win-rate) \textit{Easy} (level-2) built-in AI at about 30M frames (about 0.06M games), \textit{Hard} (level-4) at about 250M frames (about 0.5M games), \textit{VeryHard} (level-6) at about 800M frames (about 1.6M games), \textit{CheatResources} (level-9) at about 2000M (about 4M games), and \textit{CheatInsane} (level-10) at about 3500M frames (about 7M games).

After exploration and learning, the agent seems to acquire some
intriguing strategies resembling those used by human players.
We demonstrate two strategies about the combat timing (i.e., when to trigger attacks) it learns, as in Figure~\ref{fig:rl-visualize}:
\textbf{\textit{Rush}}, which triggers attacks as soon as possible, even if there are only a small number of combat units available;
\textbf{\textit{EconomyFirst}}, 
which keeps developing the economy and launches the first attack only
after having assembled a strong army. Besides, we also observed that
{\AIRL} tends to build $3\sim4$ bases to boost the economy growth and
prefers sideways when planning for the routes of attack.

\subsection{\AIHC}
Table~\ref{tab:win_rate} shows the win-rates of the agent that adopts
the hierarchical macro-micro action and rule based controllers, as
described in Section~\ref{sec:rule-bot}.
Each reported win-rate is obtained by taking the mean of 100 games with different random seeds, 
where a tie is counted as 0.5 when calculating the win-rate.
Fog-of-war is enabled during the test.
We can see that the agent is able to consistently defeat built-in AIs in all levels, 
showing the effectiveness of the hierarchical action modeling.

\subsection{TStarBots vs. Human Players}
In an informal internal test, 
we let \AIRL~or \AIHC~play against several human players
ranging from Platinum to Diamond level in the ranking system of SC2 Battle.net League.
The setting remains the same as the above sub-sections,
i.e., a Zerg-vs-Zerg full game on the map AbyssalReef with fog-of-war enabled.
The results are reported in Table~\ref{tab:vs_human}.
We can see that both \AIRL~and \AIHC~may defeat a Platinum (and even a Diamond) human player. 

\begin{table}[t]
  \centering
  \caption{
  \textbf{TStarBots vs. Human Players}. 
   Each entry means how many games TStarBot1/TStarBot2 wins and loses. 
   E.g., 1/2 means TStarBot ``wins 1 game and loses 2 games''. 
  }
  \label{tab:vs_human}
  \begin{tabular}{|c|c|c|c|c|c|}
    \hline
    \#win/\#loss & Platinum 1 & Diamond 1 & Diamond 2 & Diamond 3   \\
    \hline \hline
    TStarBot1      & 1/2        & 1/2       & 0/3       & 0/2   \\
    \hline
    TStarBot2      &1/2         & 1/0       & 0/3       & 0/2 \\
    \hline
  \end{tabular}
\end{table}

\subsection{\AIRL vs. \AIHC}
In another informal test,
we let the two TStarBots play against each other.
We observe that \AIRL~can always defeat \AIHC.
Inspecting the game-play, we find that TStarBot1 tends to use the \textit{Zergling Rush} strategy, 
while TStarBot2 lacks anti-rush strategy and henceforth always loses.
 
It is worthy of noting that although TStarBot1 can successfully learn and acquire strategies to defeat all the built-in AIs and TStarBot2, 
it lacks strategy diversity in order to consistently beat human players.
In the aforementioned test with human players, 
TStarBot1 will lose consistently once a human player figures out TStarBot1's preference for \textit{Zergling Rush}.
The insufficient strategy diversity might be caused by the following reasons.
1) A lack of opponent diversity. Although the built-in AI is already
equipped with several pre-defined strategies, their strategies are
significantly more restricted when compared to those of human players.
2) A lack of deep exploration. The production of advanced units are buried down very deeply in the tech tree, 
which is difficult and inefficient for simple exploration methods such as
epsilon-greedy to discover.
Self-play training and randomization techniques~\cite{openai-five}
are promising approaches to alleviate these issues, which we shall leave for future work.

\section{Conclusion}
\label{sec:conclusion}
For SC2LE,
we have modeled the structural action space by hand-tuned rules, 
which reduces the large number of atomic actions into a small number of
macro actions that become tractable to machine learning.
We studied two agents. 
One agent \AIRL, which is based on flat action modeling with a
reinforcement learning controller, can achieve a reasonably high
win-rate against the built-in AIs.
The  other agent \AIHC, which adopts a hierarchical action modeling and rule
based controllers, can consistently win against the built-in AIs.
This work shows that this approach of action space modeling can be
used with a
conventional RL algorithm, so that the RL algorithm can learn a good policy for the SC2 1v1 full game.
In the future, it will be beneficial to explore a unified approach,
where the hierarchy can be optimized either jointly or separately for
each component.

\subsubsection*{Acknowledgement}
It is grateful that
our colleagues Yan Wang and Lei Jiang, 
and a volunteer Yijun Huang participate the user study to test our AI agents.

\small

\bibliography{cite}

\begin{thebibliography}{10}

\bibitem{scii-forum}
Starcraft ii forum.
\newblock \url{https://eu.battle.net/forums/en/sc2/topic/853485381}.
\newblock Accessed August 30, 2018.

\bibitem{liquid}
Liquipedia - ranking system.
\newblock \url{https://liquipedia.net/starcraft2/Battle.net_Leagues}.
\newblock Accessed August 30, 2018.

\bibitem{goodfellow2016deep}
Ian Goodfellow, Yoshua Bengio, Aaron Courville, and Yoshua Bengio.
\newblock {\em Deep learning}, volume~1.
\newblock MIT press Cambridge, 2016.

\bibitem{sutton1998introduction}
Richard~S Sutton and Andrew~G Barto.
\newblock {\em Introduction to reinforcement learning}, volume 135.
\newblock MIT press Cambridge, 1998.

\bibitem{silver2016mastering}
David Silver, Aja Huang, Chris~J Maddison, Arthur Guez, Laurent Sifre, George
  Van Den~Driessche, Julian Schrittwieser, Ioannis Antonoglou, Veda
  Panneershelvam, Marc Lanctot, et~al.
\newblock Mastering the game of go with deep neural networks and tree search.
\newblock {\em nature}, 529(7587):484, 2016.

\bibitem{silver2017mastering}
David Silver, Julian Schrittwieser, Karen Simonyan, Ioannis Antonoglou, Aja
  Huang, Arthur Guez, Thomas Hubert, Lucas Baker, Matthew Lai, Adrian Bolton,
  et~al.
\newblock Mastering the game of go without human knowledge.
\newblock {\em Nature}, 550(7676):354, 2017.

\bibitem{mnih2015human}
Volodymyr Mnih, Koray Kavukcuoglu, David Silver, Andrei~A Rusu, Joel Veness,
  Marc~G Bellemare, Alex Graves, Martin Riedmiller, Andreas~K Fidjeland, Georg
  Ostrovski, et~al.
\newblock Human-level control through deep reinforcement learning.
\newblock {\em Nature}, 518(7540):529, 2015.

\bibitem{kempka2016vizdoom}
Micha{\l} Kempka, Marek Wydmuch, Grzegorz Runc, Jakub Toczek, and Wojciech
  Ja{\'s}kowski.
\newblock Vizdoom: A doom-based ai research platform for visual reinforcement
  learning.
\newblock {\em arXiv preprint arXiv:1605.02097}, 2016.

\bibitem{wu2017}
Yuxin Wu and Yuandong Tian.
\newblock Training agent for first-person shooter game with actor-critic
  curriculum learning.
\newblock In {\em International Conference on Learning Representations}, 2017.

\bibitem{beattie2016deepmind}
Charles Beattie, Joel~Z Leibo, Denis Teplyashin, Tom Ward, Marcus Wainwright,
  Heinrich K{\"u}ttler, Andrew Lefrancq, Simon Green, V{\'\i}ctor Vald{\'e}s,
  Amir Sadik, et~al.
\newblock Deepmind lab.
\newblock {\em arXiv preprint arXiv:1612.03801}, 2016.

\bibitem{openai-five}
Open ai five.
\newblock \url{https://blog.openai.com/openai-five/}.
\newblock Accessed August 30, 2018.

\bibitem{levine2016end}
Sergey Levine, Chelsea Finn, Trevor Darrell, and Pieter Abbeel.
\newblock End-to-end training of deep visuomotor policies.
\newblock {\em The Journal of Machine Learning Research}, 17(1):1334--1373,
  2016.

\bibitem{zhu2016target}
Yuke Zhu, Roozbeh Mottaghi, Eric Kolve, Joseph~J Lim, Abhinav Gupta,
  Li~Fei-Fei, and Ali Farhadi.
\newblock Target-driven visual navigation in indoor scenes using deep
  reinforcement learning.
\newblock {\em International Conference on Robotics and Automation}, 2017.

\bibitem{sadeghi2016cad}
Fereshteh Sadeghi and Sergey Levine.
\newblock (cad)2rl: Real single-image flight without a single real image.
\newblock {\em arXiv preprint arXiv:1611.04201}, 2016.

\bibitem{sc2le}
Oriol Vinyals, Timo Ewalds, Sergey Bartunov, Petko Georgiev, Alexander~Sasha
  Vezhnevets, Michelle Yeo, Alireza Makhzani, Heinrich K{\"u}ttler, John
  Agapiou, Julian Schrittwieser, et~al.
\newblock Starcraft ii: a new challenge for reinforcement learning.
\newblock {\em arXiv preprint arXiv:1708.04782}, 2017.

\bibitem{zambaldi2018relational}
Vinicius Zambaldi, David Raposo, Adam Santoro, Victor Bapst, Yujia Li, Igor
  Babuschkin, Karl Tuyls, David Reichert, Timothy Lillicrap, Edward Lockhart,
  et~al.
\newblock Relational deep reinforcement learning.
\newblock {\em arXiv preprint arXiv:1806.01830}, 2018.

\bibitem{tstarbots}
Tstarbots.
\newblock \url{https://github.com/Tencent/TStarBots}.

\bibitem{ontanon2013survey}
Santiago Ontan{\'o}n, Gabriel Synnaeve, Alberto Uriarte, Florian Richoux, David
  Churchill, and Mike Preuss.
\newblock A survey of real-time strategy game ai research and competition in
  starcraft.
\newblock {\em IEEE Transactions on Computational Intelligence and AI in
  games}, 5(4):293--311, 2013.

\bibitem{churchill2016heuristic}
David Churchill.
\newblock {\em Heuristic Search Techniques for Real-Time Strategy Games}.
\newblock PhD thesis, PhD thesis, University of Alberta, 2016.

\bibitem{tan1993multi}
Ming Tan.
\newblock Multi-agent reinforcement learning: Independent vs. cooperative
  agents.
\newblock In {\em In Proceedings of the Tenth International Conference on
  Machine Learning}, page 330–337, 1993.

\bibitem{foerster2017counterfactual}
Jakob Foerster, Gregory Farquhar, Triantafyllos Afouras, Nantas Nardelli, and
  Shimon Whiteson.
\newblock Counterfactual multi-agent policy gradients.
\newblock {\em arXiv preprint arXiv:1705.08926}, 2017.

\bibitem{usunier2016episodic}
Nicolas Usunier, Gabriel Synnaeve, Zeming Lin, and Soumith Chintala.
\newblock Episodic exploration for deep deterministic policies: An application
  to starcraft micromanagement tasks.
\newblock {\em arXiv preprint arXiv:1609.02993}, 2016.

\bibitem{peng2017multiagent}
Peng Peng, Ying Wen, Yaodong Yang, Quan Yuan, Zhenkun Tang, Haitao Long, and
  Jun Wang.
\newblock Multiagent bidirectionally-coordinated nets: Emergence of human-level
  coordination in learning to play starcraft combat games.
\newblock {\em arXiv preprint arXiv:1703.10069}, 2017.

\bibitem{sukhbaatar2016learning}
Sainbayar Sukhbaatar, Rob Fergus, et~al.
\newblock Learning multiagent communication with backpropagation.
\newblock In {\em Advances in Neural Information Processing Systems}, pages
  2244--2252, 2016.

\bibitem{mnih2016asynchronous}
Volodymyr Mnih, Adria~Puigdomenech Badia, Mehdi Mirza, Alex Graves, Timothy
  Lillicrap, Tim Harley, David Silver, and Koray Kavukcuoglu.
\newblock Asynchronous methods for deep reinforcement learning.
\newblock In {\em International conference on machine learning}, pages
  1928--1937, 2016.

\bibitem{russell2016artificial}
Stuart~J Russell and Peter Norvig.
\newblock {\em Artificial intelligence: a modern approach}.
\newblock Malaysia; Pearson Education Limited,, 2016.

\bibitem{millington2009artificial}
Ian Millington and John Funge.
\newblock {\em Artificial intelligence for games}.
\newblock CRC Press, 2009.

\bibitem{marzinotto2014towards}
Alejandro Marzinotto, Michele Colledanchise, Christian Smith, and Petter Ogren.
\newblock Towards a unified behavior trees framework for robot control.
\newblock In {\em Robotics and Automation (ICRA), 2014 IEEE International
  Conference on}, pages 5420--5427. IEEE, 2014.

\bibitem{perez2011evolving}
Diego Perez, Miguel Nicolau, Michael O’Neill, and Anthony Brabazon.
\newblock Evolving behaviour trees for the mario ai competition using
  grammatical evolution.
\newblock In {\em European Conference on the Applications of Evolutionary
  Computation}, pages 123--132. Springer, 2011.

\bibitem{tian2017elf}
Yuandong Tian, Qucheng Gong, Wenling Shang, Yuxin Wu, and C.~Lawrence Zitnick.
\newblock Elf: An extensive, lightweight and flexible research platform for
  real-time strategy games.
\newblock In I.~Guyon, U.~V. Luxburg, S.~Bengio, H.~Wallach, R.~Fergus,
  S.~Vishwanathan, and R.~Garnett, editors, {\em Advances in Neural Information
  Processing Systems 30}, pages 2659--2669. Curran Associates, Inc., 2017.

\bibitem{vezhnevets2017feudal}
Alexander~Sasha Vezhnevets, Simon Osindero, Tom Schaul, Nicolas Heess, Max
  Jaderberg, David Silver, and Koray Kavukcuoglu.
\newblock Feudal networks for hierarchical reinforcement learning.
\newblock In {\em International Conference on Machine Learning}, pages
  3540--3549, 2017.

\bibitem{van2017hybrid}
Harm Van~Seijen, Mehdi Fatemi, Joshua Romoff, Romain Laroche, Tavian Barnes,
  and Jeffrey Tsang.
\newblock Hybrid reward architecture for reinforcement learning.
\newblock In {\em Advances in Neural Information Processing Systems}, pages
  5392--5402, 2017.

\bibitem{sutton1999between}
Richard~S Sutton, Doina Precup, and Satinder Singh.
\newblock Between mdps and semi-mdps: A framework for temporal abstraction in
  reinforcement learning.
\newblock {\em Artificial intelligence}, 112(1-2):181--211, 1999.

\bibitem{bacon2017option}
Pierre-Luc Bacon, Jean Harb, and Doina Precup.
\newblock The option-critic architecture.
\newblock In {\em AAAI}, pages 1726--1734, 2017.

\bibitem{ghazanfari2017autonomous}
Behzad Ghazanfari and Matthew~E Taylor.
\newblock Autonomous extracting a hierarchical structure of tasks in
  reinforcement learning and multi-task reinforcement learning.
\newblock {\em arXiv preprint arXiv:1709.04579}, 2017.

\bibitem{vezhnevets2016strategic}
Alexander Vezhnevets, Volodymyr Mnih, Simon Osindero, Alex Graves, Oriol
  Vinyals, John Agapiou, et~al.
\newblock Strategic attentive writer for learning macro-actions.
\newblock In {\em Advances in neural information processing systems}, pages
  3486--3494, 2016.

\bibitem{frans2017meta}
Kevin Frans, Jonathan Ho, Xi~Chen, Pieter Abbeel, and John Schulman.
\newblock Meta learning shared hierarchies.
\newblock {\em arXiv preprint arXiv:1710.09767}, 2017.

\bibitem{van2016deep}
Hado Van~Hasselt, Arthur Guez, and David Silver.
\newblock Deep reinforcement learning with double q-learning.
\newblock In {\em AAAI}, volume~2, page~5. Phoenix, AZ, 2016.

\bibitem{wang2016dueling}
Ziyu Wang, Tom Schaul, Matteo Hessel, Hado Hasselt, Marc Lanctot, and Nando
  Freitas.
\newblock Dueling network architectures for deep reinforcement learning.
\newblock In Maria~Florina Balcan and Kilian~Q. Weinberger, editors, {\em
  Proceedings of The 33rd International Conference on Machine Learning},
  volume~48 of {\em Proceedings of Machine Learning Research}, pages
  1995--2003, New York, New York, USA, 20--22 Jun 2016. PMLR.

\bibitem{schulman2017proximal}
John Schulman, Filip Wolski, Prafulla Dhariwal, Alec Radford, and Oleg Klimov.
\newblock Proximal policy optimization algorithms.
\newblock {\em arXiv preprint arXiv:1707.06347}, 2017.

\bibitem{bellemare2016unifying}
Marc Bellemare, Sriram Srinivasan, Georg Ostrovski, Tom Schaul, David Saxton,
  and Remi Munos.
\newblock Unifying count-based exploration and intrinsic motivation.
\newblock In {\em Advances in Neural Information Processing Systems}, pages
  1471--1479, 2016.

\bibitem{schulman2015trust}
John Schulman, Sergey Levine, Pieter Abbeel, Michael Jordan, and Philipp
  Moritz.
\newblock Trust region policy optimization.
\newblock In {\em International Conference on Machine Learning}, pages
  1889--1897, 2015.

\bibitem{schulman2015high}
John Schulman, Philipp Moritz, Sergey Levine, Michael Jordan, and Pieter
  Abbeel.
\newblock High-dimensional continuous control using generalized advantage
  estimation.
\newblock {\em arXiv preprint arXiv:1506.02438}, 2015.

\end{thebibliography}
\bibliographystyle{unsrt}

\newpage
\section*{Appendix I: List of Macro Actions}
\vskip -0.35in
\begin{table}[h!]
  \centering
  \small
  \caption{List of all macro actions (for \textit{Zerg} race only)}
  \label{tab:macro_action_list}
  \begin{tabular}{|c|l|}
    \hline
    Categories & Macro Actions \\
    \hline \hline
    \multirow{13}{*}{Building} & \textit{BuildExtractor} \\
    \cline{2-2} & \textit{BuildSpawningPool} \\
    \cline{2-2} & \textit{BuildRoachWarren} \\
    \cline{2-2} & \textit{BuildHydraliskDen} \\
    \cline{2-2} & \textit{BuildHatchery} \\
    \cline{2-2} & \textit{BuildEvolutionChamber} \\
    \cline{2-2} & \textit{BuildBanelingNest} \\
    \cline{2-2} & \textit{BuildInfestationPit} \\
    \cline{2-2} & \textit{BuildSpire} \\
    \cline{2-2} & \textit{BuildUltraliskCaven} \\
    \cline{2-2} & \textit{BuildNydusNetwork} \\
    \cline{2-2} & \textit{BuildSpineCrawler} \\
    \hline
    \multirow{22}{*}{Production} & \textit{ProduceDrone} \\
    \cline{2-2} & \textit{ProduceZergling} \\
    \cline{2-2} & \textit{ProduceRoach} \\
    \cline{2-2} & \textit{ProduceHydralisk} \\
    \cline{2-2} & \textit{ProduceViper} \\
    \cline{2-2} & \textit{ProduceMutalisk} \\
    \cline{2-2} & \textit{ProduceCorruptor} \\
    \cline{2-2} & \textit{ProduceSwarmHost} \\
    \cline{2-2} & \textit{ProduceInfestor} \\
    \cline{2-2} & \textit{ProduceUltralisk} \\
    \cline{2-2} & \textit{ProduceOverlord} \\
    \cline{2-2} & \textit{ProduceQueen} \\
    \cline{2-2} & \textit{ProduceNydusWorm} \\
    \cline{2-2} & \textit{MorphLurkerDen} \\
    \cline{2-2} & \textit{MorphLair} \\
    \cline{2-2} & \textit{MorphHive} \\
    \cline{2-2} & \textit{MorphGreaterSpire} \\
    \cline{2-2} & \textit{MorphBaneling} \\
    \cline{2-2} & \textit{MorphRavager} \\
    \cline{2-2} & \textit{MorphLurker} \\
    \cline{2-2} & \textit{MorphBroodlord} \\
    \cline{2-2} & \textit{MorphOverseer} \\
    \hline
  \end{tabular}
  \begin{tabular}{|c|l|}
    \hline
    Categories & Macro Actions \\
    \hline \hline
    \multirow{27}{*}{Upgrading} & \textit{UpgradeBurrow} \\
    \cline{2-2} & \textit{UpgradeCentrificalHooks} \\
    \cline{2-2} & \textit{UpgradeChitionsPlating} \\
    \cline{2-2} & \textit{UpgradeEvolveGroovedSpines} \\
    \cline{2-2} & \textit{UpgradeEvolveMuscularAugments} \\
    \cline{2-2} & \textit{UpgradeGliareConstituion} \\
    \cline{2-2} & \textit{UpgradeInfestorEvergy} \\
    \cline{2-2} & \textit{UpgradeNeuralParasite} \\
    \cline{2-2} & \textit{UpgradeOverlordSpeed} \\
    \cline{2-2} & \textit{UpgradeTunnelingClaws} \\
    \cline{2-2} & \textit{UpgradeFlyerArmorsLevel-1} \\
    \cline{2-2} & \textit{UpgradeFlyerArmorsLevel-2} \\
    \cline{2-2} & \textit{UpgradeFlyerArmorsLevel-3} \\
    \cline{2-2} & \textit{UpgradeFlyerWeaponLevel-1} \\
    \cline{2-2} & \textit{UpgradeFlyerWeaponLevel-2} \\
    \cline{2-2} & \textit{UpgradeFlyerWeaponLevel-3} \\
    \cline{2-2} & \textit{UpgradeGroundArmorsLevel-1} \\
    \cline{2-2} & \textit{UpgradeGroundArmorsLevel-2} \\
    \cline{2-2} & \textit{UpgradeGroundArmorsLevel-3} \\
    \cline{2-2} & \textit{UpgradeZerglingAttackSpeed} \\
    \cline{2-2} & \textit{UpgradeZerglingMoveSpeed} \\
    \cline{2-2} & \textit{UpgradeMeleeWeaponsLevel-1} \\
    \cline{2-2} & \textit{UpgradeMeleeWeaponsLevel-2} \\
    \cline{2-2} & \textit{UpgradeMeleeWeaponsLevel-3} \\
    \cline{2-2} & \textit{UpgradeMissileWeaponsLevel-1} \\
    \cline{2-2} & \textit{UpgradeMissileWeaponsLevel-2} \\
    \cline{2-2} & \textit{UpgradeMissileWeaponsLevel-3} \\
    \hline
    \multirow{3}{*}{Harvesting} & \textit{CollectMinerals} \\
    \cline{2-2} & \textit{CollectGas} \\
    \cline{2-2} & \textit{InjectLarvas} \\
    \hline
    \multirow{4}{*}{Combating} & \textit{ZoneA-Attack-ZoneB} \\
    \cline{2-2} & \textit{ZoneA-Attack-ZoneC} \\
    \cline{2-2} & ...... \\
    \cline{2-2} & \textit{ZoneJ-Attack-ZoneJ} \\
    \hline
  \end{tabular}
\end{table}

\end{document}